\documentclass[letterpaper, 10 pt, conference]{ieeeconf}  

\IEEEoverridecommandlockouts                              

\usepackage{graphics} 
\usepackage{epsfig} 
\usepackage{times} 
\usepackage{amsmath} 
\usepackage{amssymb}  
\usepackage{cite}
\usepackage{subcaption}
\usepackage[normalem]{ulem}
\usepackage[ruled,vlined,linesnumbered]{algorithm2e}
\usepackage{xargs}    
\usepackage{textcomp}
\usepackage{multirow}
\usepackage{lipsum}
\usepackage{subcaption}
\usepackage{xcolor}
\usepackage{bm}
\usepackage{float}
\usepackage{comment}
\usepackage{todonotes}

\title{\LARGE \bf
Domain Curiosity: Learning Efficient Data Collection Strategies for Domain Adaptation
}

\author{Karol Arndt$^{1}$, Oliver Struckmeier$^{1}$, and Ville Kyrki$^{1}$
\thanks{This work was supported by Academy of Finland grant 317020. We acknowledge the computational resources provided by the Aalto Science-IT project.
}
\thanks{$^{1}$Intelligent Robotics Group, Aalto University, Espoo, Finland
        {\tt\small first.last@aalto.fi}}%
}

\begin{document}

\newcommand{\change}[1]{\textcolor[rgb]{1,0,0}{#1}}

\maketitle
\thispagestyle{empty}
\pagestyle{empty}

\hyphenation{spots-and-honey-pots}
\begin{abstract}
Domain adaptation is a common problem in robotics, with applications such as transferring policies from simulation to real world and lifelong learning.
Performing such adaptation, however, requires informative data about the environment to be available during the adaptation.
In this paper, we present \textit{domain curiosity}---a method of training exploratory policies that are explicitly optimized to provide data that allows a model to learn about the unknown aspects of the environment. 
In contrast to most curiosity methods, our approach explicitly rewards learning, which makes it robust to environment noise without sacrificing its ability to learn.
We evaluate the proposed method by comparing how much a model can learn about environment dynamics given data collected by the proposed approach, compared to standard curious and random policies.
The evaluation is performed using a toy environment, two simulated robot setups, and on a real-world haptic exploration task.
The results show that the proposed method allows data-efficient and accurate estimation of dynamics.
\end{abstract}

\section{Introduction}
Since the advent of reinforcement learning methods capable of handling complex, continuous state spaces~\cite{silver14ddpg,Haarnoja18soft,schulamn2015trpo,schulman2017proximal}, the ability to train control policies using only data coming from interacting with the environment has gained interest in the robotics community.
Modern reinforcement learning methods, however, still require extensive interactions with the environment before learning useful behaviours, which may take from hours~\cite{Haarnoja18soft} to months~\cite{andrychowicz19learning} of real-time interaction. 
To reduce the dependence on large amounts of real-world data, a policy is often trained in a simulated environment, where the general structure of the problem is learned, and then deployed in the physical world.
This scenario is known as \textit{domain transfer}.

Domain transfer in reinforcement learning has been extensively studied, especially in the context of sim-to-real transfer in robotics~\cite{andrychowicz19learning,valassakis2020crossing,transferMurtaza}.
Multiple approaches to this problem have been proposed in the past, such as training a robust policy that performs well across a range of dynamics conditions~\cite{murattore19assessing,chebotar19closing}, training a policy conditioned on the belief over dynamics~\cite{zintgraf2019varibad}, or training a policy that is able to adapt to new dynamic conditions~\cite{Nagabandi2019learning,arndt2019meta}.
However, robust policies do not generalize well when the possible range of dynamics is wide~\cite{mehta20active}, especially in situations where the task requires high precision or the optimal actions depend heavily on the dynamics~\cite{valassakis2020crossing}. 

Adaptation and domain identification approaches, on the other hand, use data from the target environment; this dataset has to be complex enough to allow the model to identify the dynamics parameters, and yet small enough to be feasible to collect on physical hardware within a reasonable time.
The data collection process also needs to be safe to perform on physical hardware which rules out random policies. In addition, dynamics identification requires the actions to be identifiable which poses a problem e.g.\ with kinesthetic human demonstrations, where only states may be directly measured. Gathering the data for dynamics adaptation is thus a central problem for domain adaptation and identification.

In this paper, we propose the \textit{domain curiosity} approach to learning a policy that efficiently collects such adaptation data.
In contrast to previously proposed intrinsic rewards in the domain adaptation context, our method rewards the agent for providing data that allows a dynamics model to improve its predictions about the environment, instead of rewarding prediction error.
This is achieved by jointly training a meta-learned adaptable dynamics model together with an exploration policy.

We experimentally evaluate the method on a toy environment and on two simulated robotics setups: a sliding task under unknown friction and a haptic exploration task, where the robot's goal is to locate the position of an object in the scene.
We also deploy the policy for the latter setup on a real-world Franka Panda robot arm.

\begin{figure}
    \centering
    \begin{subfigure}{0.49\linewidth}
    \includegraphics[width=\linewidth]{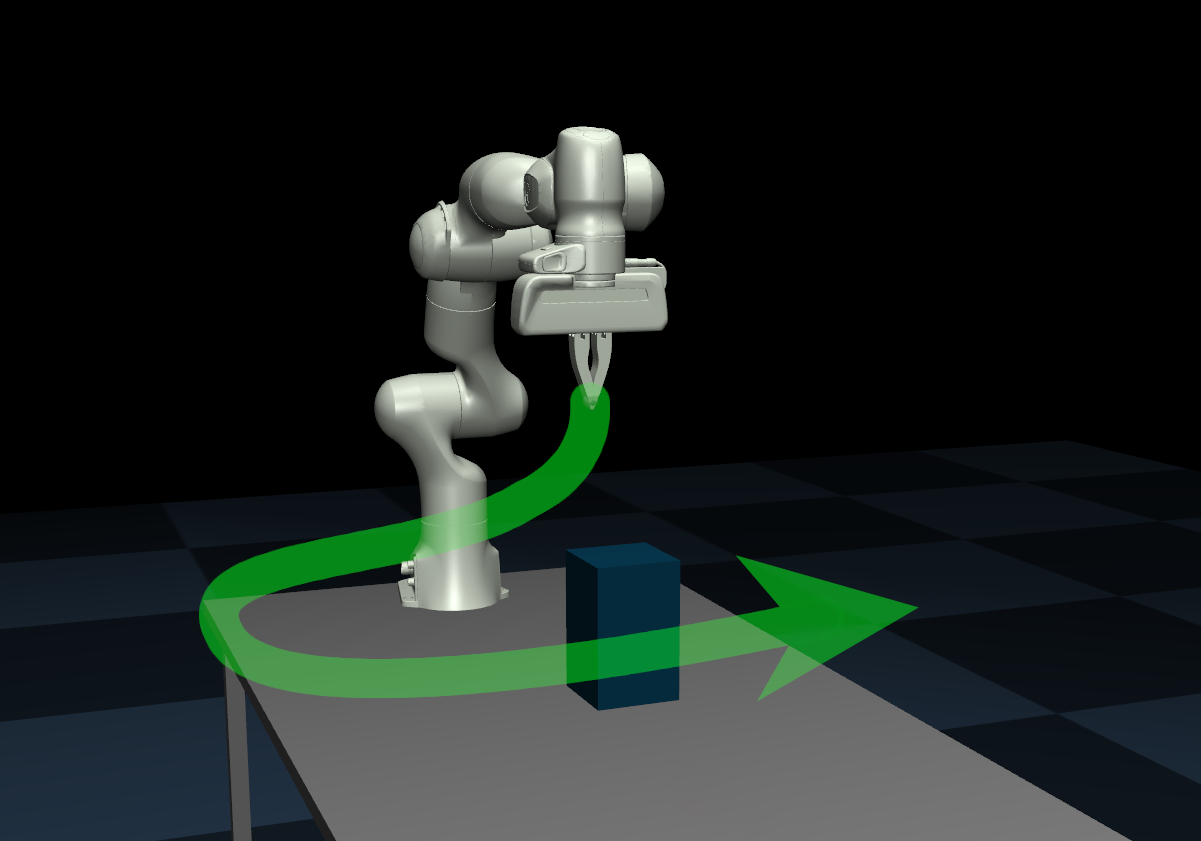}
    \caption{}
    \label{fig:intro_box}
    \end{subfigure}
    \begin{subfigure}{0.49\linewidth}
    \includegraphics[width=\linewidth]{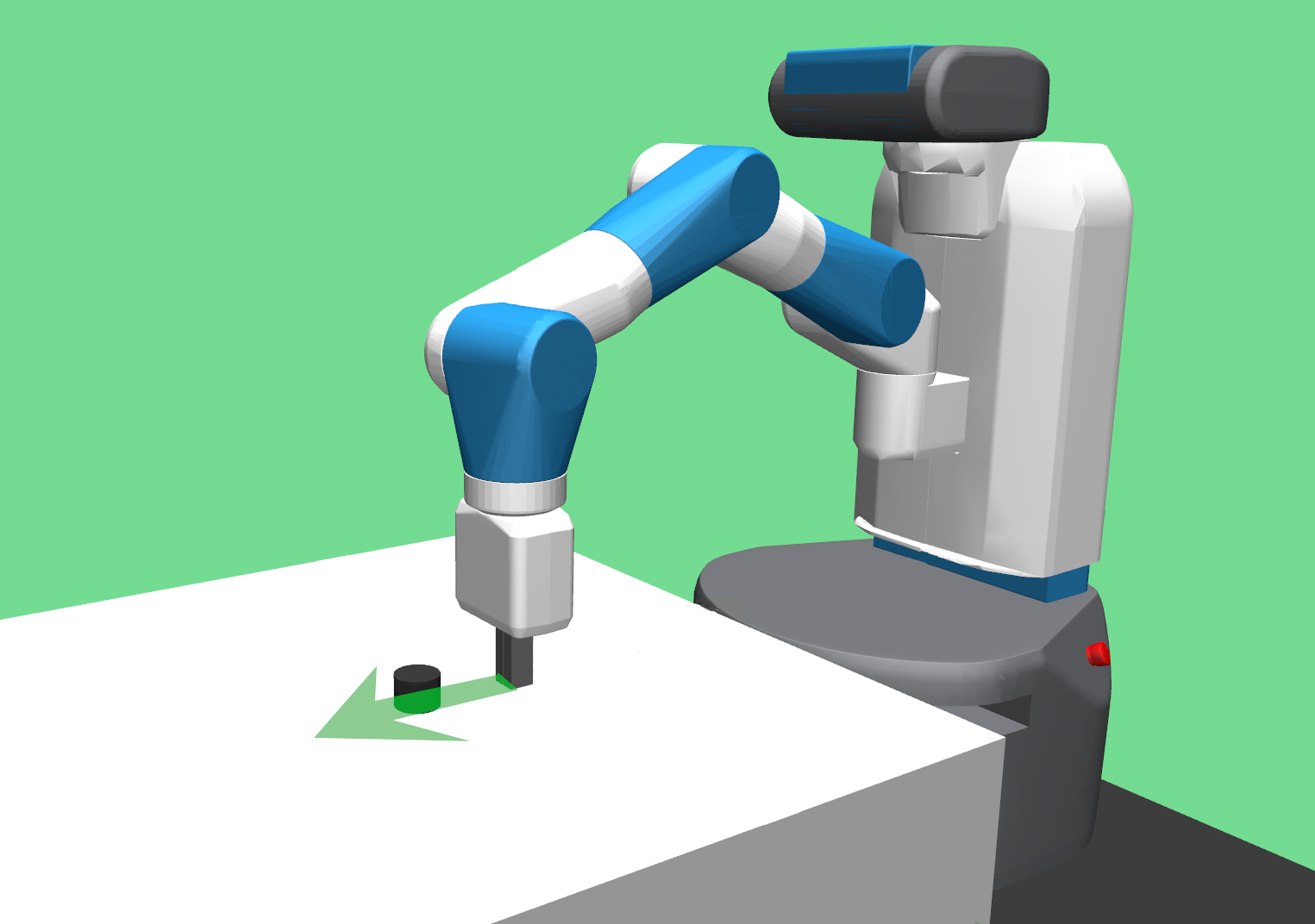}
    \caption{}
    \label{fig:intro_fetch}
    \end{subfigure}
    \caption{The proposed method learns efficient strategies to collect data for dynamics estimation, as shown in a haptic exploration setup (\subref{fig:intro_box}) and on a sliding setup with unknown friction parameters (\subref{fig:intro_fetch}).}
    \label{fig:my_label}
\vspace{-2em}
\end{figure}

The contributions of this paper are (1) proposing a novel framework for training a curious policy that maximizes the agent's information gain about the domain-specific aspects the environment, (2) providing experimental evaluation in simulated and real world environments, (3) showing that the proposed domain curious approach is able to reliably provide data which helps to identify the dynamic conditions in the environment and to perform better next state prediction.

\section{Related work}
\vspace{-2pt}
\subsection{Curiosity in reinforcement learning}
\vspace{-2pt}
The idea of intrinsic, curiosity-based exploration was first described by Schmidhuber~\cite{schmidhuber1991possibility}.
In order to focus on epistemic (\textit{i.e.,} learnable) uncertainty, this method was later modified to reward model improvement~\cite{Schmidhuber91adaptiveconfidence,Schmidhuber2010FormalTO}, model parameter updates~\cite{houthooft2016vime}, work in feature space~\cite{pathak2017curiosity} and reward disagreement between ensemble dynamics models~\cite{pathak19disagreement,shyam19modelbased}, instead of operating on raw state predictions.

Our method also learns to focus on epistemic uncertainty by rewarding model improvement, similarly to~\cite{Schmidhuber91adaptiveconfidence}; however, our method is designed for a domain adaptation scenario, which allows it to distinguish prediction errors induced by lack of knowledge about the domain from pure noise.

Our method shares the basic theoretical formulation with VIME~\cite{houthooft2016vime}.
VIME maximizes the information gain of an agent about a distribution of dynamics models in addition to the extrinsic reward from the environment.
The information gain is measured using variational inference with a reward term based on the KL divergence between two following hidden states of a model.
This method, however, considers only a single training domain, while our approach combines it with the meta-learning formulation and applies it to the multi-domain reinforcement learning scenario.

\vspace{-2pt}
\subsection{Exploration in meta-reinforcement learning}
\vspace{-2pt}
In meta-RL, various ways of learning exploration strategies have been previously explored.
The original model-agnostic meta-learning (MAML;~\cite{finn2017maml}) formulation only optimizes for post-adaptation performance, without explicitly optimizing the meta-policy to provide useful samples for adaptation.
Despite this shortcoming, MAML has been successfully applied to domain adaptation and sim-to-real transfer~\cite{arndt2019meta,Nagabandi2019learning}.
This problem was later addressed in~\cite{stadie2018emaml} and~\cite{rothfuss2018promp}, where the agent was additionally incentivised to provide useful trajectories during adaptation.
This resulted in more stable and more consistent training procedure.

More recently, VariBAD~\cite{zintgraf2019varibad} was shown to perform close to Bayes-optimal exploration in simple gridworlds by performing approximate variational inference over a hidden parameter and conditioning a dynamics model on that hidden parameter.
The method implicitly learns to explore by using this hidden parameter to meta-learn a policy on an unknown task based on previous experience.
However, it does not directly include the prediction accuracy in the reward function and requires an extrinsic task reward.
In contrast, our method directly encourages collecting useful data by rewarding the correction to the model's predictions.

Zhang \textit{et al.}~\cite{zhang2020learn} described a meta-learning framework based on similar grounding to VIME; however, in the practical implementation, this approach was simplified to embedding a basic surprise-based curiosity reward inside a meta-reinforcement learning agent.

A limitation of previous meta-RL methods is the dependence on on-policy data.
Rakelly \textit{et al.}~\cite{rakelly2019efficient} address this with a stochastic component performing inference over a latent context variable to encourage exploration and ultimately faster adaption to new tasks.
Domain curiosity on the other hand jointly learns the task and a dynamics model in a way that promotes exploration of informative components of the environment to achieve the same effect.

\vspace{-2pt}
\section{Method}
\vspace{-2pt}
\subsection{Definitions and problem formulation}
\vspace{-2pt}
The domain adaptation problem problem can be formulated as a partially observable Markov decision process (MDP) of the form $\mathcal{M}=(\mathcal{S}, \mathcal{A}, p_{tr}, p_{s_0}, r, \xi)$,
where $\mathcal{S}$ is the state space, $\mathcal{A}$ is the action space, $\xi$ is a vector of dynamics parameters, $p(s_0)$ is the initial state probability distribution, $p_{tr}: \mathcal{S} \times \mathcal{A} \times \mathcal{S} \times \Xi \rightarrow \mathcal{R}$ represents the transition probabilities conditioned on $\xi$, and $r: \mathcal{S} \times \mathcal{A} \times \mathcal{S} \rightarrow \mathcal{R}$ is the reward function.
By following a policy $\pi(a|s)$, the agent collects trajectories $\tau$, where each trajectory is a sequence of states and actions: $\tau = (s_0, a_0, s_1, a_1, ..., a_{T-1}, s_T)$.
These trajectories form the dataset: $\mathcal{D} = \{\tau_0, ..., \tau_N\}$.

Then, the specific problem addressed in this paper can be formulated as follows: find a policy $\pi(a|s)$ that collects a trajectory dataset $\mathcal{D}$ that maximize the agent's information gain about the dynamics parameters $\xi$.
More specifically, knowing $\xi$ reduces the entropy of dynamics transitions, $\mathcal{H}(p_{tr}(s, a, s' | \xi)) \leq \mathcal{H}(p_{tr}(s, a, s'))$.
Since each individual state transition depends on $\xi$, so does $\mathcal{D}$: $p(\mathcal{D}|\xi)$, and thus $\mathcal{D}$ can, in turn, be used to estimate $\xi$: $p(\xi | \mathcal{D})$.
All in all, the agent's information gain can be expressed by the entropy difference $\mathcal{H}(p_{tr}(s, a, s')) - \mathcal{H}(p_{tr}(s, a, s'| \mathcal{D}))$.

\vspace{-2pt}
\subsection{Policy training}
\vspace{-2pt}
\begin{figure*}
    \centering
    \includegraphics[width=.9\linewidth]{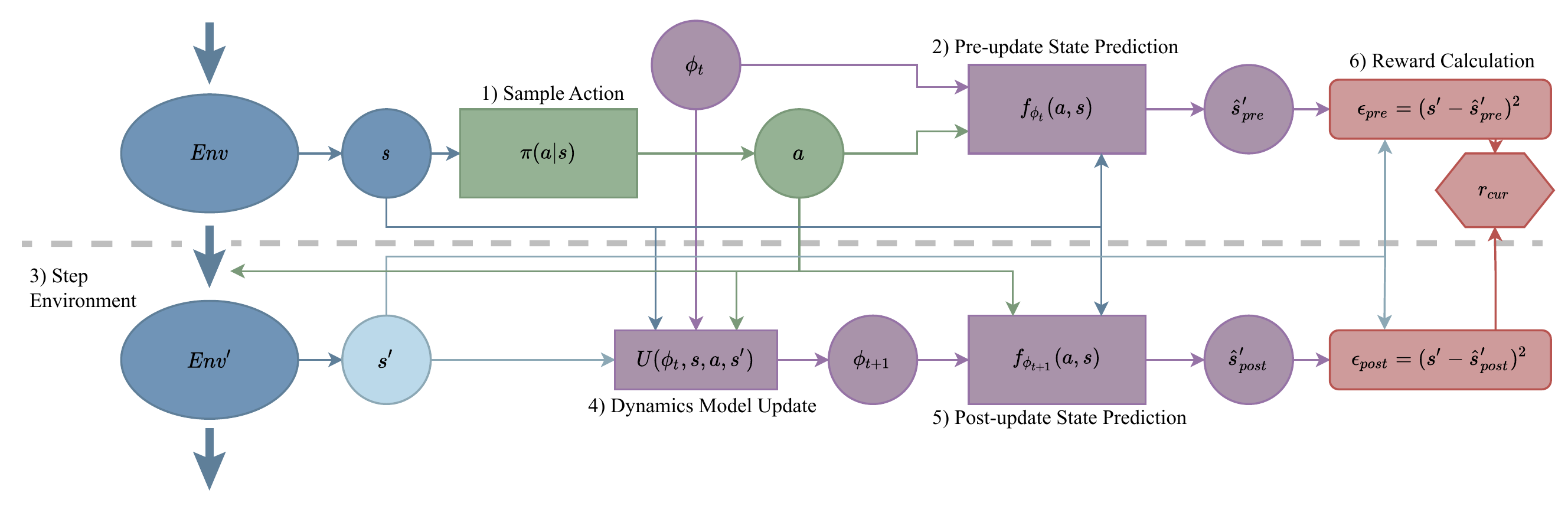}
    
    \vspace{-12pt}
    \caption{Overview of the proposed reward calculation method. The elements of the agent are marked in green, the environment---in blue, and the dynamics model in purple.}
    \label{fig:method}
    \vspace{-18pt}
\end{figure*}

The training procedure for Domain curiosity is outlined in detail in Algorithm~\ref{alg:domcur}, and an overview of the reward computation procedure is shown in Figure~\ref{fig:method}.
The main idea is to have an adaptable dynamics model of the environment $f_\phi(s, a)$, whose prediction updates can be used as a basis for reward computation.
These rewards are then used to train an exploratory policy.
Our method assumes that the dynamics model is not available in advance, and trains one from scratch using data collected by the policy.

The algorithm is initialized by randomly setting the policy weights $\theta$ and the model weights $\phi$, and initializing two empty buffers for storing state transitions: one for the policy $\mathcal{D}_\pi$ and one for the dynamics model $\mathcal{D}_f$ (steps~\ref{as:init_p}--\ref{as:init_d}).
On-policy methods like PPO~\cite{schulman2017proximal} require $\mathcal{D}_\pi$ to be cleared after every policy update; since model training can be performed off-policy, using a single, shared buffer in this context would be wasteful, and could make model training unstable due to data distribution shift.

Each training iteration starts by sampling a random task $\xi$, which---in the domain adaptation context---defines the dynamics of the system.
At the same time we also reset the environment (step~\ref{as:reset_env}) and set the belief over the hidden state back to the prior (step~\ref{as:reset_params}).

\begin{algorithm}
\SetAlgoLined  
\KwResult{Parameters $\theta$ of a policy that identifies $\xi$}
 randomly initialize $\theta$ and $\phi_0$ \; \label{as:init_p}
 initialize empty replay buffers $\mathcal{D}_\pi$ and $\mathcal{D}_f$ \; \label{as:init_d}
 \While{not done}{
  Sample a random task $\xi_i \sim p(\xi)$\; \label{as:sample}
  Reset the env; $s \sim p(s_0)$\; \label{as:reset_env}
  Reset the model parameters; $\phi_t = \phi_0$\; \label{as:reset_params}
 \While{episode not over}{
   Sample $a \sim \pi_\theta(a|s)$\; \label{as:sample_a}
   Calculate pre-update state prediction (Eq. \ref{eq:pre_upd_pred}) \;  \label{as:pre_upd}
   Execute $a$, observe $s'$\; \label{as:exec_a}
   Update the dynamics model (Eq. \ref{eq:update}) \;  \label{as:update}
   Calculate post-update state prediction (Eq. \ref{eq:post_upd_pred}) \;  \label{as:post_upd}
   Calculate the reward (Eqs. \ref{eq:cur_rew} and \ref{eq:rew}) \; \label{as:rew}
   Store $(s, a, r, s')$ in $\mathcal{D}_{\pi}$ and $\mathcal{D}_{f}$ \; \label{as:store}
 }
 Update $\pi_\theta$ with $\mathcal{D}_{\pi}$ to maximize expected cumulative return\;
 Update $\phi_0$ with $\mathcal{D}_{f}$ to minimize prediction errors over each sequence (Eq.~\ref{eq:model_loss})\;
 }
 \caption{Domain curiosity}
 \label{alg:domcur}
\end{algorithm}

For each episode, we follow the policy $\pi_\theta$.
We first sample an action from the policy (step~\ref{as:sample_a} in Alg.~\ref{alg:domcur} and step 1 in Fig.~\ref{fig:method}).
We then use the current dynamics model to predict the next state (step~\ref{as:pre_upd} in Alg.~\ref{alg:domcur} and step 2 in Fig.~\ref{fig:method}) as 
\begin{equation} \label{eq:pre_upd_pred}
    \hat{s}'_{pre} = f_{\phi_t}(s, a).
\end{equation}

Next, we step the environment using $a$ and observe the true next state $s'$ (step~\ref{as:exec_a} in Algorithm~\ref{alg:domcur} and step 3 in Figure~\ref{fig:method}).
Then, we update the belief over dynamics using the observed transition (step~\ref{as:update} in Alg.~\ref{alg:domcur} and step 4 in Fig.~\ref{fig:method}) by
\begin{equation} \label{eq:update}
\vspace{-1pt}
    \phi_{t+1} = U(\phi_t, s, a, s'),
\vspace{-1pt}
\end{equation}
where $U(\cdot)$ represents the inner-loop update rule used by the dynamics model (e.g., in case of MAML, $U$ would correspond to performing a step of gradient descent).

To quantify the \textit{improvement} of the dynamics model after observing the transition, we predict the post update state $\hat{s}'_{post}$ using the dynamics model with the \textit{updated} state belief from Eq. \ref{eq:update} (step~\ref{as:post_upd} in Alg.~\ref{alg:domcur} and step 5 in Fig.~\ref{fig:method}):
\begin{equation} \label{eq:post_upd_pred}
\vspace{-1pt}
    \hat{s}'_{post} = f_{\phi_{t+1}}(s, a).
\vspace{-1pt}
\end{equation}
This effectively evaluates if observing the transition allows the model to make more accurate prediction about it.

While it may seem that having observed the transition would always allow the model to make a good prediction by simply memorizing the value that has just been passed to it, this would only be the case with a model that memorizes the observed samples instead of generally modelling the environment.
This was an important consideration for selecting the architecture of the dynamics model; we provide more details behind it in Section~\ref{sec:method_model}.
To provide a tangible example, observing a sample from the unit Gaussian distribution does not allow one to make more accurate predictions about future samples from that distribution (assuming the distribution that the sample came from is known).
This step allows our method to tell apart prediction errors coming from noise (\textit{aleatoric uncertainty}) and prediction errors coming from insufficient knowledge of the domain the model has been deployed in (\textit{epistemic uncertainty}).
It also differentiates the proposed method from standard curiosity-based intrinsic rewards.

We finally calculate the prediction error for both methods $\varepsilon_{pre} = (s' - \hat{s}'_{pre})^2$ and $\varepsilon_{post} = (s' - \hat{s}'_{post})^2$.
The domain curiosity reward term is then calculated as the different between the two prediction errors, corresponding to the reduction in prediction error (step~\ref{as:rew} in Alg.~\ref{alg:domcur} and step 6 in Fig.~\ref{fig:method})
\begin{equation} \label{eq:cur_rew}
\vspace{-1pt}
    r_{cur} = \varepsilon_{pre} - \varepsilon_{post}
\vspace{-1pt}
\end{equation}

To regularize the policy and to make it physically feasible, we also add a control cost penalty
\begin{equation} \label{eq:rew}
\vspace{-1pt}
    r = r_{control} + \lambda r_{cur}
\vspace{-1pt}
\end{equation}
The control cost, expressed by the reward term $r_{control}$, penalizes large control forces and---in case of robot environments---large contact forces between the robot and the objects in the environment. The $\lambda$ hyperparameter is used to balance between control cost and the curiosity reward.

We then store the observed transition $(s, a, s')$ in $\mathcal{D}_f$ in order to update the dynamics model, and $(s, a, r, s')$ in $\mathcal{D}_\pi$ for training the policy.
After finishing each episode, we update the policy $\pi_\theta$ and the dynamics model $f_\phi$.
The policy update can be performed with an arbitrary reinforcement learning algorithm.

\vspace{-1pt}
\subsection{Dynamics model} \label{sec:method_model}
\vspace{-1pt}
Generally speaking, the proposed policy training approach can be used with an arbitrary adaptable meta-learned dynamics model.
We found MAML~\cite{finn2017maml}to be impractical due to the large computational cost of performing an update every timestep over a long episode, and due to the potential of memorizing the last-seen sample, which is facilitated through the gradient-descent inner-loop learning rule.
We also found the classic LSTM approach~\cite{hochreiter2001learning} to be unsuitable, as the post-update state prediction step in the proposed method requires the calculation of output given the updated hidden state ($f(x_t, h_t)$), whereas, during training, the model is never trained to make such predictions (only $f(x_t, h_{t-1})$); such samples also cannot be directly included in the training procedures, as that creates a direct path in the model between the true value $y_t$ and the model's prediction $\hat{y}$, which encourages the model to "cheat" by simply memorizing the previous observed sample in $h_t$.
This would effectively cause the model to make accurate post-update predictions of observation and/or system noise.

Given these considerations, we selected the backprop Kalman filter (BKF)-based approach proposed in~\cite{arndt2020fewshot}.
The model is separated into three parts: dynamics measurement, belief update, and next state prediction.
Unlike the recurrent method in~\cite{hochreiter2001learning}, the architecture is constructed such that the recurrent hidden state is optimized to only encode information about the dynamics and does not need to store information about previously observed samples.
This allows the reward calculation procedure to query arbitrary dynamics beliefs with arbitrary state-action pairs.
In this formulation, model parameter vector can be split in two parts: the \textit{static} parameters, which are only updated in the outer loop (all the neural network weights) and the task specific parameters, updated by $U(\cdot)$ (the hidden state of the filter, as defined by the mean and variance of the state belief).

To further facilitate learning over long sequences, we change the loss function to calculate be the sum of the prediction loss over the whole sequence:

\vspace{-6pt}
\begin{equation} \label{eq:model_loss}
    \mathcal{L} = \sum_{t=0}^T \sum_{n=0}^N (s'_n - f_{\phi_t}(s_n, a_n))^2
\end{equation}
\vspace{-6pt}

With this change, it is important to keep the data used to compute the values of $\phi_t$ separate from the state transitions used to calculate the loss (in contrast to~\cite{hochreiter2001learning}, where the same data is used for both tasks, but the target labels are shown to the model only after it has already made the prediction for their corresponding input).
This prevents the model from memorizing the observed samples in the hidden state.
While we find it unlikely to occur in practice, it would theoretically still be possible if the selected dimensionality of the hidden states $\phi$ were too large. 
We also found it beneficial to add L2 norm regularization on $\phi$, in the form of adding $\sum_{t=0}^T \phi_t^2$ to the loss term.

\vspace{-8pt}

\vspace{-1pt}
\section{Experiments}
\vspace{-1pt}
In this section, we present the results of the experimental evaluation in the toy environment, in two simulated robotics setups and on a real-world setup with the Franka Panda robot.

In all experiments, the policy is trained using Proximal Policy Optimization (PPO; ~\cite{schulman2017proximal}), but---in general---the method can be used with an arbitrary reinforcement learning algorithm.
During the first updates, the model's predictions and inner-loop updates are random, which can have an adverse effect on learning.
To reduce this effect and, effectively, stabilize the training, we only update the dynamics model in the early phases of training, and later gradually start training the policy by linearly increasing the PPO clip coefficient from 0 to the final value.
We perform 4 epochs of model updates per each policy update and use a model replay buffer of 25000 episodes in all experiments.
The $\lambda$ hyperparameter, which balances the curious reward and the control cost, is set to $10^4$ in all environments.

\vspace{-3pt}
\subsection{Toy example}
\vspace{-3pt}
To demonstrate the method on a simple example and provide an evaluation with a Bayes-optimal task variable prediction model, we developed a simple environment, \textit{SpotsAndHoneypots}, shown in Figure~\ref{fig:snh}.
In this environment, an agent has to learn to explore task specific variables by visiting three \textit{spots} (marked as green, purple and orange circles in Figure~\ref{fig:snh}) that provide noisy estimates of those variables.
The environment also includes a \textit{honeypot} that acts as a source of random noise.
The state space is $(x,y,\alpha) \in R^3$ where $(x,y)$ is the position of the agent and $\alpha$ is a variable that depends on the current dynamics, $\xi$.
The action space consists of the agent's desired velocity in each direction $a = (v_x, v_y)$.
At the beginning of an episode the agent, represented by a black dot, starts in the center of the environment.

Whenever one of the spots is visited by the agent, the third element in the state vector is a noisy measurement of one of the task variables, proportionally to the distance from the center of the spot: $\alpha \sim \mathcal{N}(\mu_\alpha, 10^{-4})$, $\mu_\alpha= (1-\frac{d}{\rho}) \xi_i$, where $i$ is the index of the visited spot, $\rho$ is its radius, and $d$ is the distance to center.
Thus, in order to learn about the task-specific aspects of the environment, the agent needs to explore all three spots.

In honeypot, the yellow spot in Figure~\ref{fig:snh}, $\alpha$ is a random value sampled from a zero-mean Gaussian distribution: $\alpha \sim \mathcal{N}(0, 25)$.
Similarly to the \textit{spots}, this value is also scaled by the distance to center ($1-\frac{d}{r}$).
The task variable describing the system dynamics, $\xi$, is a three-element vector, with each element sampled independently from $\mathcal{U}(-8, 8)$.

In this environment, a standard 'greedy' curious algorithm is attracted to the yellow honeypot (as shown in Figure~\ref{fig:vc_paths}), while the proposed method learns a policy that moves around the whole environment, visiting all three spots and thus revealing information about all elements of the task variable $\xi$, as shown in Figure~\ref{fig:dc_paths}.
This result also experimentally verifies that the trained dynamics model does not memorize random noise samples in the hidden state, as described in Section~\ref{sec:method_model}---if that were the case, the model would learn to overfit to each observed sample from the honeypot, which have a larger standard deviation than the spots.
A random agent (Figure~\ref{fig:random5_paths}), with actions sampled from a normal distribution, also sometimes visits the spots, and is thus capable of collecting some information about the environment, but is far less efficient than the domain curious approach.

\subsubsection{Bayes-optimal model} \label{sec:stan}
The simple nature of the environment makes it feasible to quantitatively evaluate the quality of the data provided by each method, by evaluating how much a Bayes-optimal model can learn about the task variable $\xi$ with data provided by each method. 
To do this, we implemented the model in Stan~\cite{carpenter2017stan}, a Bayesian inference framework.
We evaluate each model by measuring the log-probability of the true value of $\xi$ under the $\xi$ distribution inferred by the model $n$ timesteps into the episode, with $n \in \{10, 20, 50, 100, 200, 500\}$.

\begin{figure}
    \centering
    \begin{subfigure}{0.45\linewidth}
    \centering
    \includegraphics[width=\linewidth]{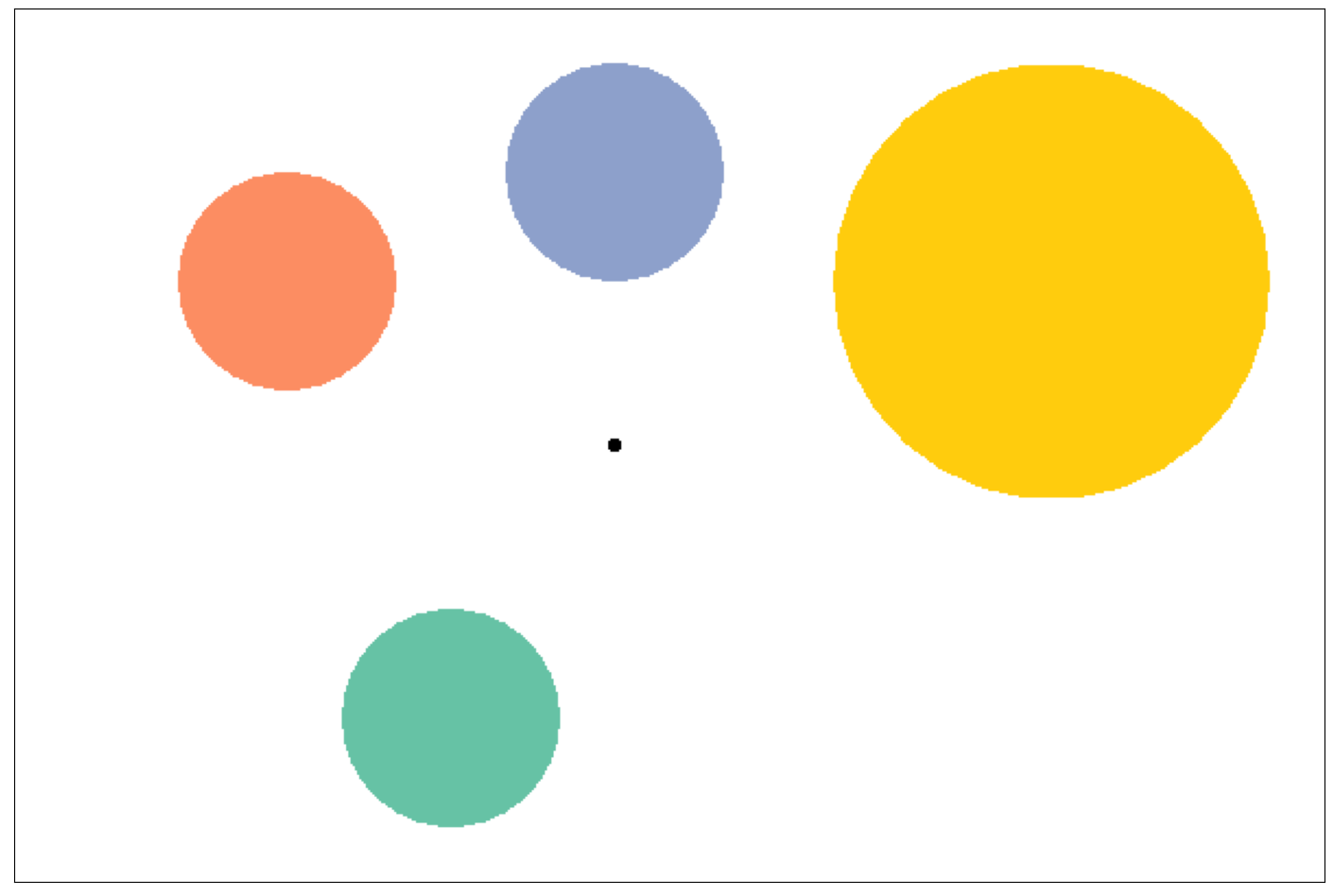}
    \caption{}
    \label{fig:snh}
    \end{subfigure}
    \begin{subfigure}{0.45\linewidth}
    \centering
    \includegraphics[width=\linewidth]{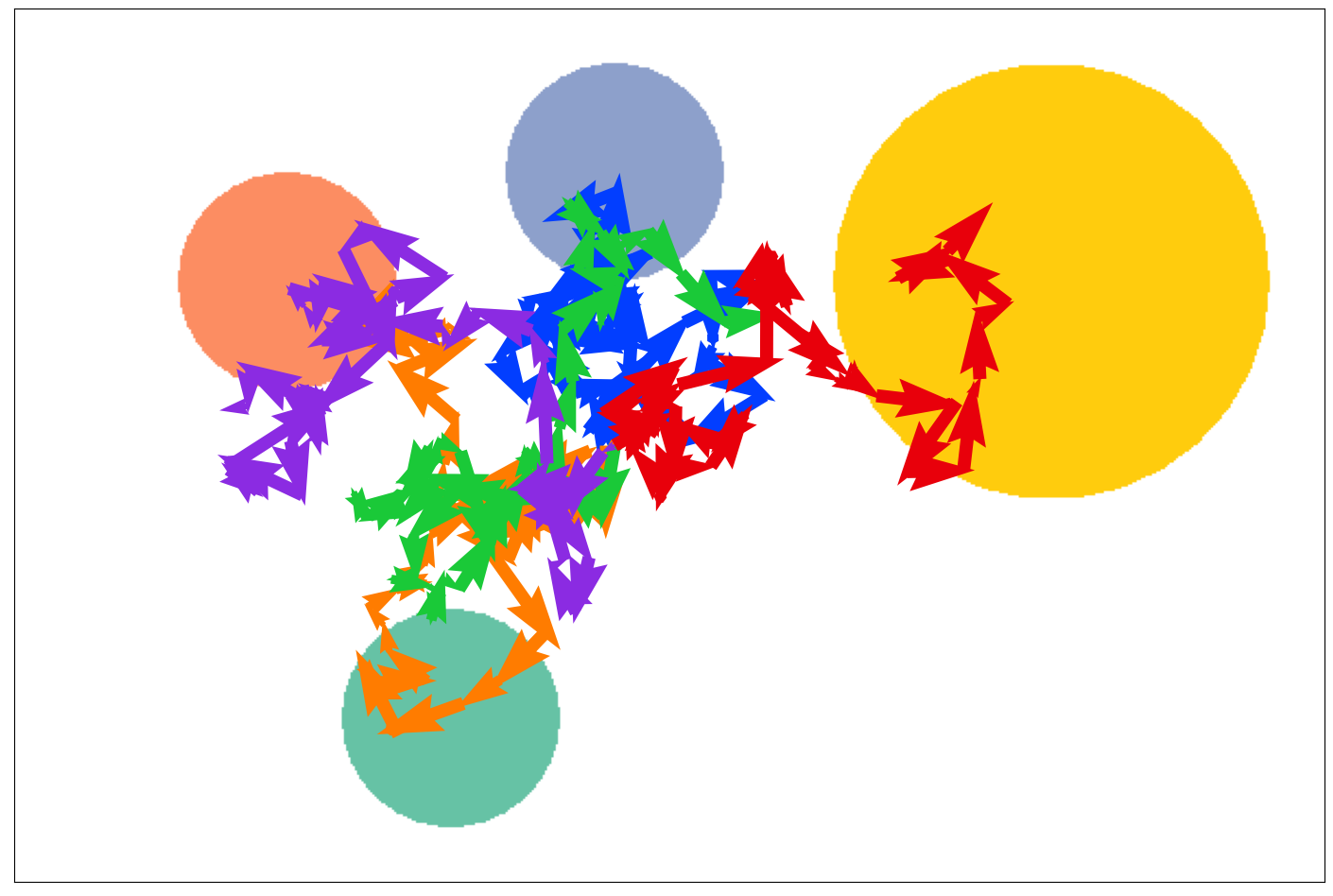}
    \caption{}
    \label{fig:random5_paths}
    \end{subfigure}
    \begin{subfigure}{0.45\linewidth}
    \centering
    \includegraphics[width=\linewidth]{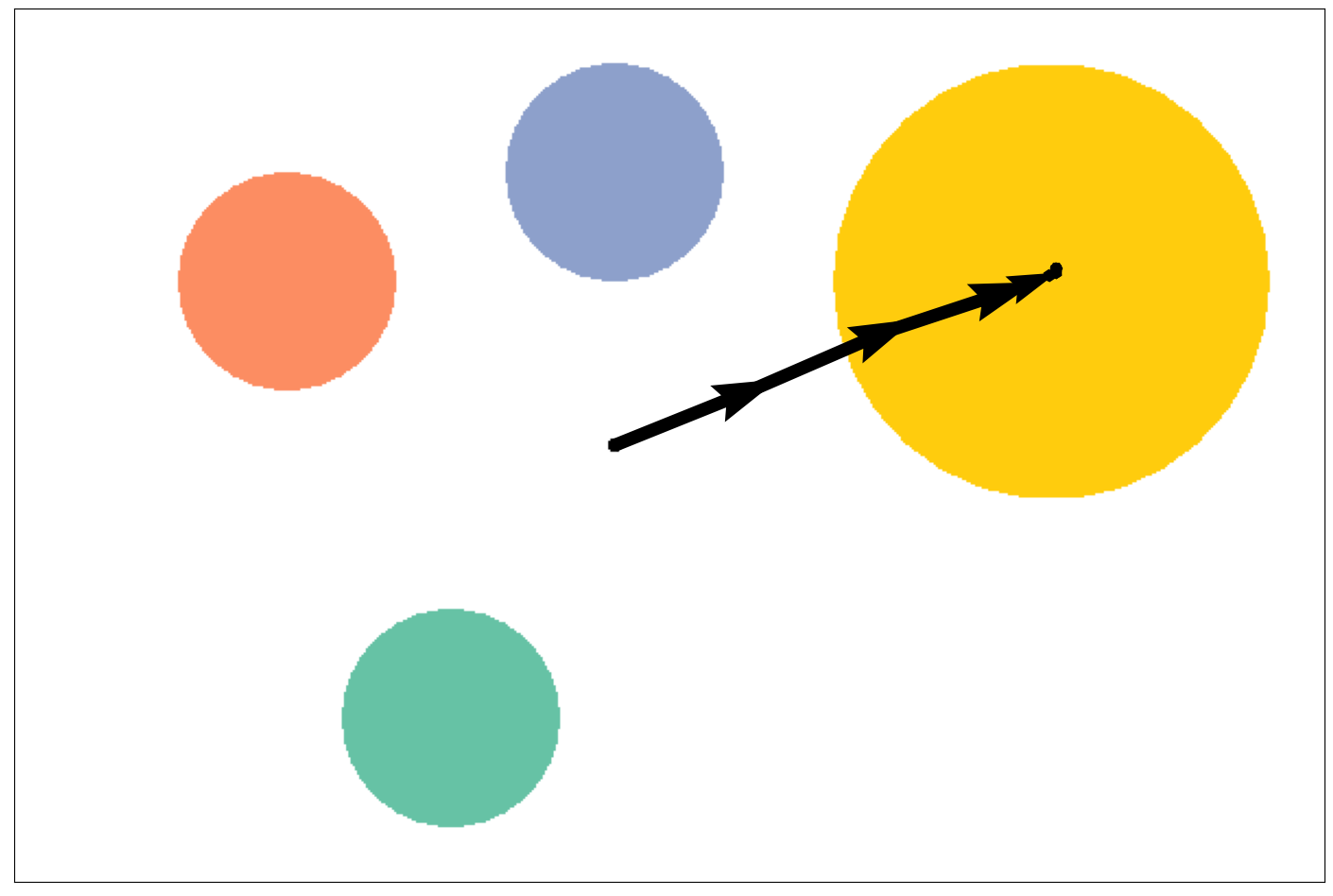}
    \caption{}
    \label{fig:vc_paths}
    \end{subfigure}
    \begin{subfigure}{0.45\linewidth}
    \centering
    \includegraphics[width=\linewidth]{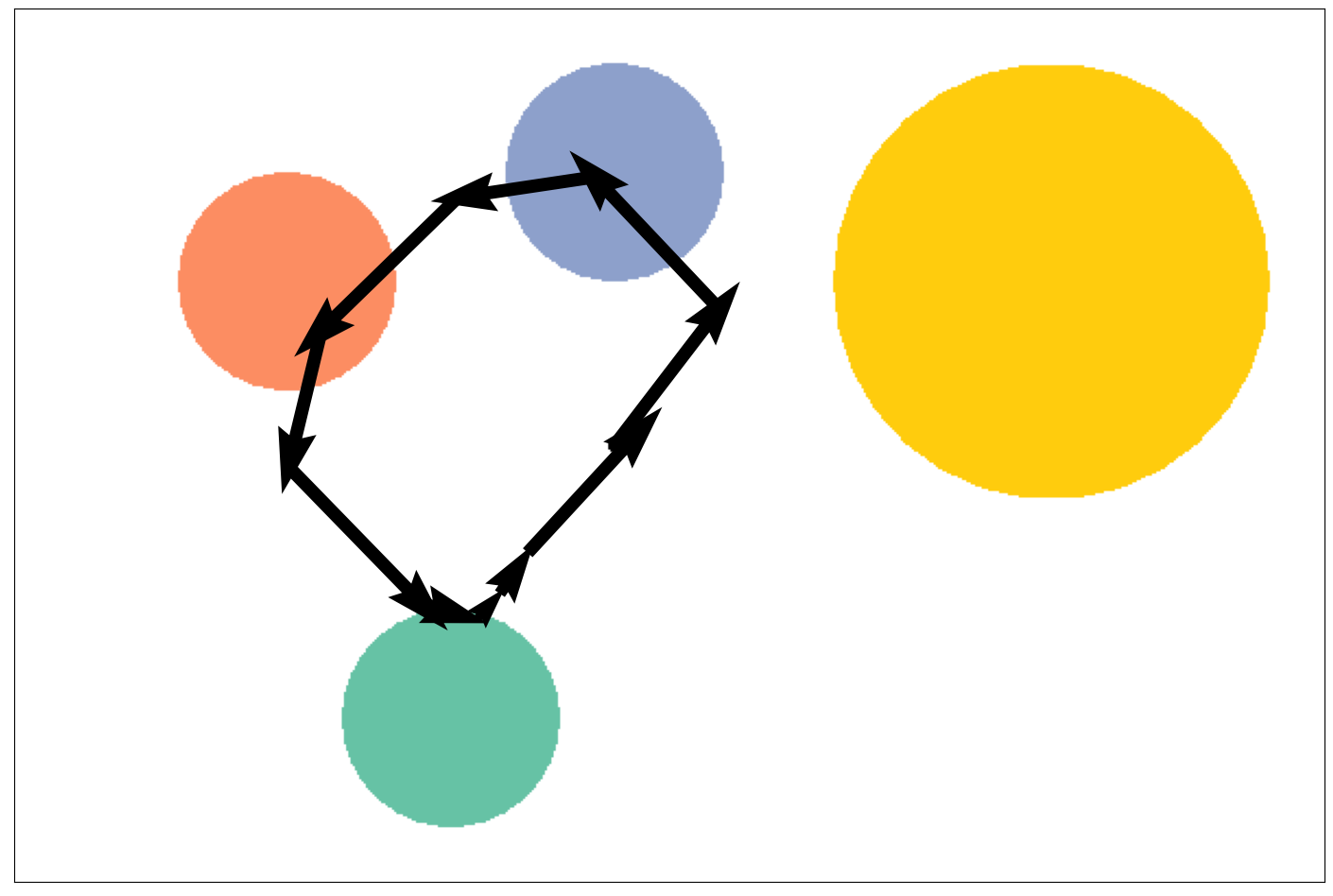}
    \caption{}
    \label{fig:dc_paths}
    \end{subfigure}
    \vspace{-6pt}
    \caption{The SpotsAndHoneypots environment (\subref{fig:snh}) and trajectories taken by each policy during exploration: 5 rolluouts of a random policy (\subref{fig:random5_paths}), a standard curious policy (\subref{fig:vc_paths}), and the proposed method (\subref{fig:dc_paths}).}
    \label{fig:snh}
    \vspace{-8pt}
\end{figure}

\begin{figure}
    \centering
    \includegraphics[width=.8\linewidth]{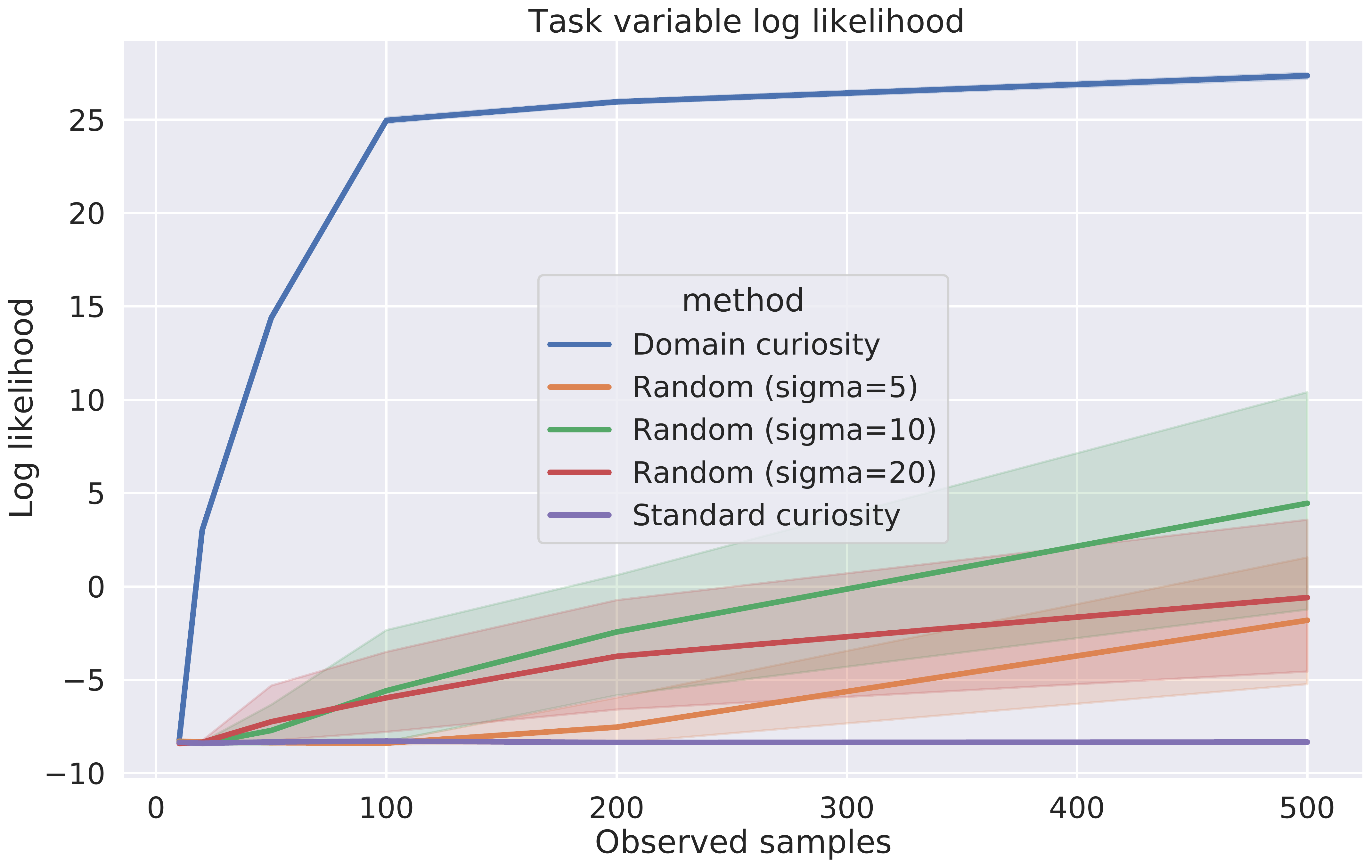}
    \vspace{-4pt}
    \caption{Task variable log likelihood in SpotsAndHoneypots, as inferred by a Bayes-optimal model with data collected by each method}
    \label{fig:stan_res}
    \vspace{-20pt}
\end{figure}

The results of this evaluation are shown in Figure~\ref{fig:stan_res}.
We observe that the domain curious approach provides data that allows to identify the task variable accurately and with high confidence, compared to other approaches.
A standard curious policy, which learns to go to the honeypot (as shown in Figure~\ref{fig:vc_paths}), never collects any useful data about the environment, as it is attracted by environment that is highly unpredictable.
A random policy gains more and more confidence as it keeps exploring, but does not get close to the results achieved by the proposed domain curiosity approach within a single episode.
Notably, a random policy with standard deviation of 10 performs the best out of all tested values.
Smaller values, such as 5, tend to stay close to the center or explore only one spot, whereas  larger values, such as 20, tend to get stuck along the edges of the environment; a value of 10 provides a tradeoff between both behaviors.

\begin{table}[]
    \centering
    \begin{tabular}{c||c|c|c}
        Method & SpotsAndHoneypots & FetchSlide & PandaFindBox \\
        \hline
        Random           & 6.3   & 0.092   & 0.21 \\
        Task policy      & --    & 0.053   & -- \\
        Random reach     & --    & --      & 0.20 \\
        Curiosity        & 7.7   & 0.044   & 0.21 \\
        Uniform          & 2.6    & --      & -- \\
        Domain curiosity & \textbf{0.7} & 0.042 & \textbf{0.03}
    \end{tabular}
\vspace{-5pt}
    \caption{Task variable prediction error after 50 (SpotsAndHoneypots, FetchSlide) or 200 samples (PandaFindBox). }
    \label{tab:tau_predict}
    \vspace{-8pt}
\end{table}

\begin{table}[]
    \centering
    \begin{tabular}{c||c|c|c}
        Method & SpotsAndHoneypots & FetchSlide & PandaFindBox \\
        \hline
        Random          & 0.81      &   $5.18 \cdot 10^{-4}$   & $7.8 \cdot 10^{-4}$ \\
        Task policy     & --        &   $4.15 \cdot 10^{-4}$   & -- \\
        Random reach    & --        &   --   & $7.8 \cdot 10^{-4}$ \\
        Curiosity       & 0.81      &   $4.09 \cdot 10^{-4}$   & $7.8 \cdot 10^{-4}$ \\
        Uniform         & 0.77      & --   & -- \\
        Domain curiosity & 0.68     &   $4.08 \cdot 10^{-4}$   & $7.6 \cdot 10^{-4}$ 
    \end{tabular}
\vspace{-5pt}
    \caption{Mean next state prediction error after observing 50 samples using data collected by each method.}
    \label{tab:state_predict}
    \vspace{-15pt}
\end{table}

\subsubsection{$\xi$ identification}
To demonstrate how reliably the task variables can be predicted given a set of collected transitions we train a recurrent neural network to predict the value of the task variable.
The neural network used for these experiments is built up of three fully connected layers with ReLU nonlinearities, followed by an LSTM cell, and by three more fully connected layers with 64 neurons each.
The hidden state of the LSTM is 8-dimensional.

As an additional baseline, we also include the \textit{Uniform} dataset, which is collected by placing the agent in a uniform place in the environment, and taking a random action.
This dataset has no time consistency between samples and thus cannot be realistically collected by any policy; however, it does not have any bias towards the starting position, like all the datasets collected by all tested policies do.
For the random policy, in this experiment, we report values for the standard deviation of 10, as---just like in the previous experiment---it is the one that performed best out of the options we tried (1, 5, 10, and 20).

The results of this evaluation are presented in the first column of Table~\ref{tab:tau_predict}.
We see that, similarly to the results obtained from the Bayes optimal model, the domain curious model allows for the most accurate estimation of the dynamics parameters, with the Uniform data distribution ranking second, and the standard curious model performing worst (i.e, not learning anything about the environment).

\subsubsection{Next state prediction}
In addition to task variable identification, we also asses the state prediction accuracy of an adaptable dynamics model, using data collected by each of the methods.
The test data for all tests in SpotsAndHoneypots was sampled from the \textit{Uniform} data distribution; thus, in this evaluation, we assess the model's ability to learn about predicting transitions in the uniform dataset, in an off-policy fashion using data collected by each method for learning.
The model we use for this purpose is a recurrent network using the LSTM cells; we chose this architecture in order to be able to identify potential bias introduced by the BKF-based architecture used for optimizing the policy (thus, we keep the two architectures different).

The results of this evaluation are presented in the first column of Table~\ref{tab:state_predict}.
We can see that, with the proposed method, the model is capable of learning the most about the state transitions in the Uniform dataset, with a noticeably smaller prediction error than other methods.
The remaining error---0.68---can mostly be attributed to the honeypot, where the dynamics model cannot learn to make any predictions about the $\alpha$ element of the state vector.
Similarly to the $\xi$ identification task, data collected by the 'standard' curious algorithm does not provide any useful data.

The experiments in SpotsAndHoneypots show that the proposed method is capable of identifying task-specific elements of the environment, while not overfitting to observation noise.
The simple structure environment allowed us to perform evaluation using a Bayes-optimal dynamics model for $\xi$ identification, and later to approximate this procedure with a recurrent neural network.
In later experiments, we evaluate the method's capability of handling more complex robotics setups.

\vspace{-2pt}
\subsection{FetchSlide}
\vspace{-2pt}
Our first robotics setup is the FetchSlide environment from OpenAI Gym~\cite{openaigym}, where the goal is to slide an object to a target position on a surface with unknown friction.
The state space contains the positions and velocities of the robot end-effector and the puck.
The action space contains Cartesian velocity of the end-effector and a command to open or close the gripper.
To randomize the environment, we randomize the friction between the puck and the surface, similarly to~\cite{arndt2020fewshot}.

In this setup, we do not provide Bayes-optimal results, as it is infeasible to implement a physics simulation engine inside a Bayesian inference framework.

\subsubsection{Dynamics identification}
For this evaluation, we use the same method as was used for the task variable identification in SpotsAndHoneypots.
The model is trained to predict friction, given a sequence of state transitions from a single episode collected by each of the methods.
The model is regularized by adding a small Gaussian noise $\epsilon \sim \mathcal{N}(0, 4\cdot10^{-4})$ to the position and velocity measurements used as the state.
The results of this evaluation are presented in Table~\ref{tab:tau_predict}.

Based on the results, we see that both the curious and the domain curious algorithm learned to provide more or less equally informative data.
This can be attributed to the lack of randomness in the environment; essentially, the only thing the feedforward dynamics model can be inaccurate about, is the motion of the puck; thus, domain curiosity and the standard curiosity approach both learn to hit the puck in a way that allows for dynamics identification.
Interestingly, most curiosity methods aimed at aiding exploration, such as~\cite{pathak2017curiosity}, would actually learn to ignore this source of randomness, as it would be treated as environment noise and ignored by the model in the feature space; thus, for exploration tasks related to domain identification, the usual benefits of this method (apparent in single-domain exploration) would actually have an adverse effect in exploration related to domain identification.

The performance of both random policies and the standard curious algorithm in this setup is also aided by the ease of exploration in this environment---the puck is always initialized within 10cm of the position of the gripper, which makes this source of model error easy to find for the curious algorithm.

We also see that both curious methods perform slightly better than the task policy. 
This could be caused by the puck moving noticeably faster in both methods, which makes the changes between consecutive states larger, making it easier to identify changes in position and velocity, and thus improving the model's ability to estimate the dynamics.

\subsubsection{Next state prediction}
Similarly to FetchSlide, we also evaluate the model's ability to predict state transitions given data collected by each of the approaches.
The reference policy---the one whose state transitions are being predicted---is, in this case, the task policy, which slides the puck to the goal position.
In this experiment, we only predict the motion of the puck, and ignore other elements of the state space (such as joint positions), as they are not affected by the friction in any way.
Thus, in this experiment we assess how well an adaptable, LSTM-based dynamics model can predict the movement of the puck, as caused by a trained task policy, given data collected from each exploration method.

The results summarized in Tab. \ref{tab:state_predict} show that both the domain curious and standard curious approaches allow for significantly smaller prediction than a random policy, and slightly smaller than the task policy.
There are no significant differences between the two curiosity approaches.

\vspace{-1pt}
\subsection{Box}
\vspace{-1pt}
In this setup, the Franka Panda robot is tasked with exploring the environment to locate the position of a heavy box.
The position of the box is randomly selected in range of 50--70 cm along the $x$ axis (facing away from the robot) and -40--40cm along the $y$ axis (which spans the whole range between the left and right edge of the table).
The state space includes the current joint position and velocities of the arm (and does not include the position of the box).
The action space contains desired accelerations of the robot (which makes it relatively easy to enforce torque, acceleration and velocity limits of a physical robot arm).
The robot is controlled by a compliant, low-gain impedance controller.
The robot is able to identify the position of the box by coming in contact with and pushing onto it---due to large mass of the box and low gains of the impedance controller, it is difficult for the robot to move the box.
This feedback signal, as expressed by the difference between where the robot \textit{expected} to move and where it \textit{actually} moved, can be used to identify the location of the box.

\vspace{-1pt}
\subsection{Learned motions}
\begin{figure}
    \centering
    \begin{subfigure}{0.48\linewidth}
    \centering
    \includegraphics[width=\linewidth]{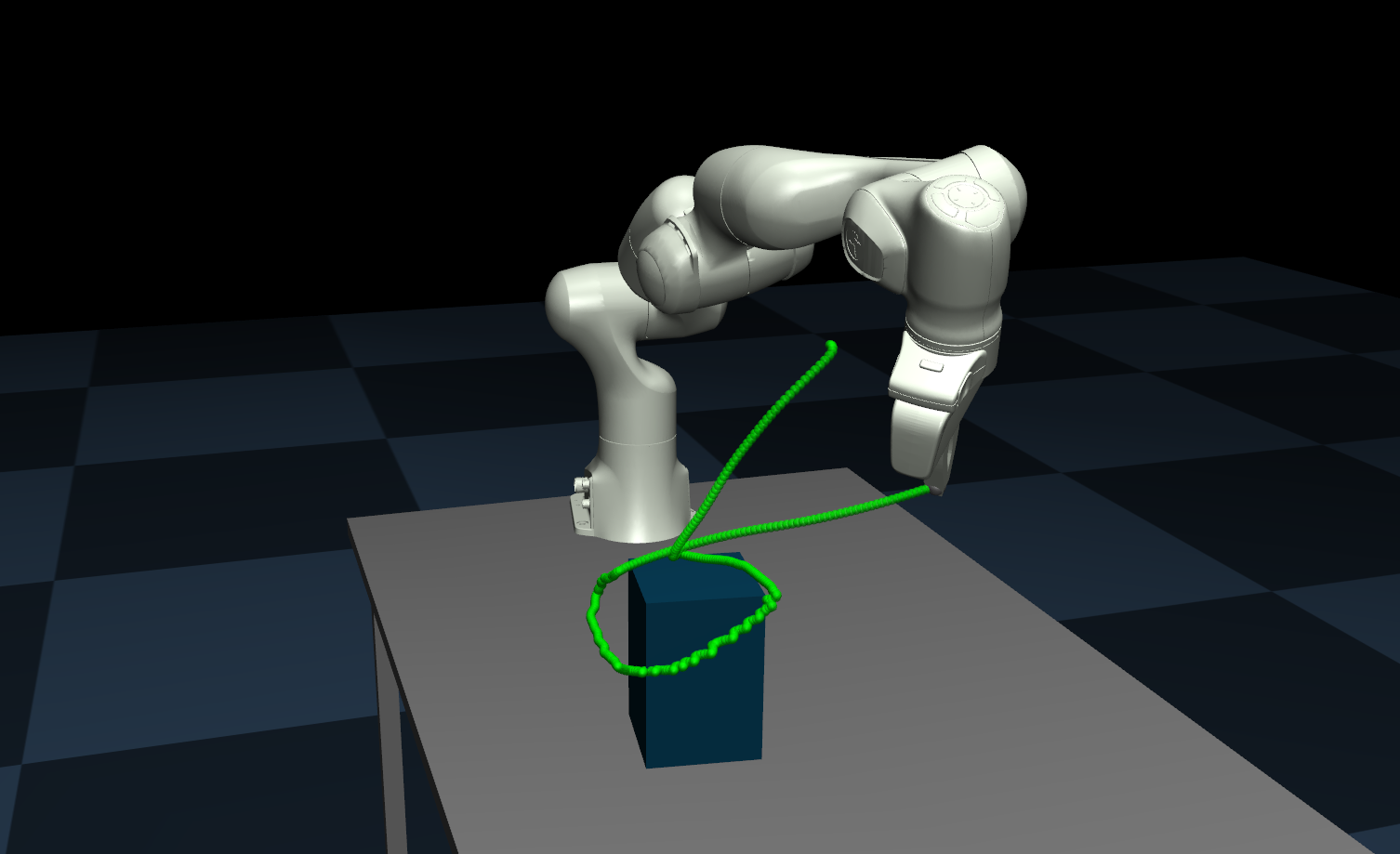}
    \caption{}
    \label{fig:traj1}
    \end{subfigure}
    \begin{subfigure}{0.48\linewidth}
    \centering
    \includegraphics[width=\linewidth]{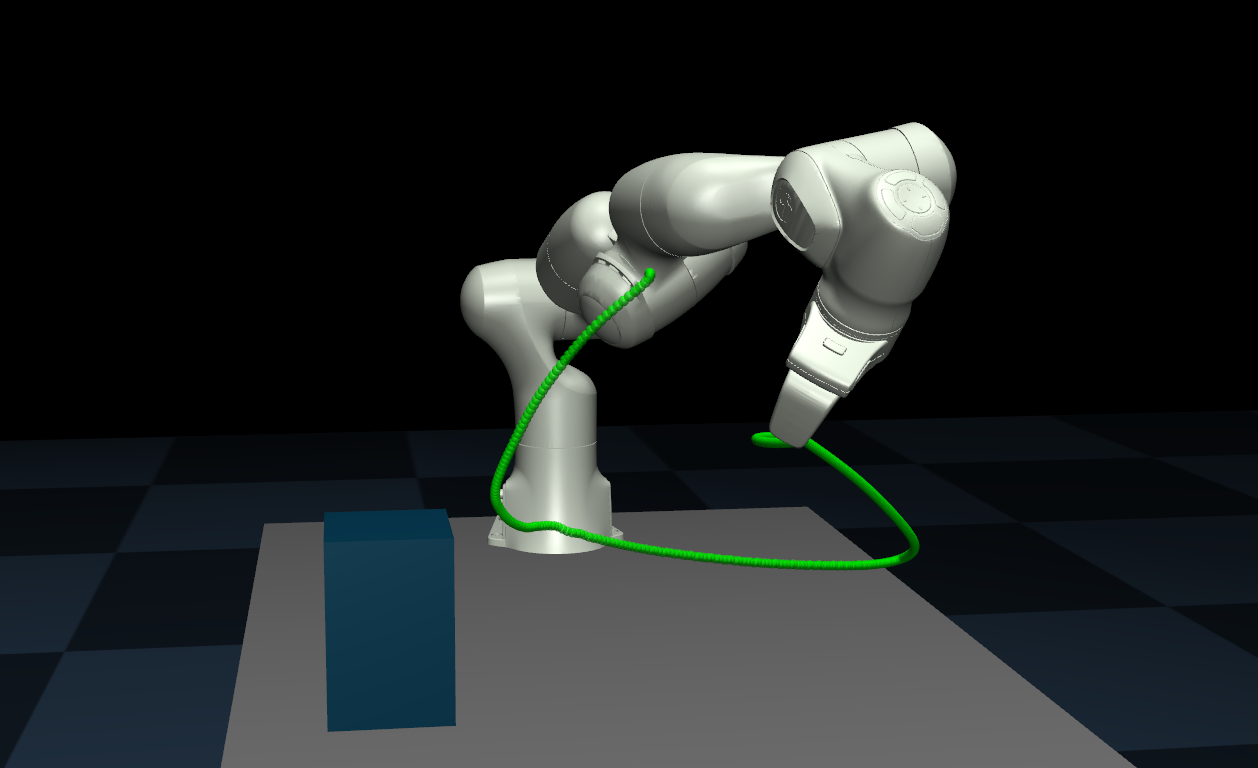}
    \caption{}
    \label{fig:traj4}
    \end{subfigure}
    \vspace{-4pt}
    \caption{Example end-effector paths (marked in green) learned by the policy in the Box environment}
    \label{fig:pandatrajs}
    \vspace{-14pt}
\end{figure}

\vspace{-1pt}
This policy learns sweeping motions, covering most of the space between the edges of the table, and trying to follow the edges of the box when it is encountered.
The specific motion depends on the random seed, with a few example paths shown in Figure~\ref{fig:pandatrajs}.
In Figure~\ref{fig:traj1}, the policy encountered the box during the sweeping motion and started to move the end-effector around it.
Figure~\ref{fig:traj4} shows a failure case, where the box is located too close to the edge of the table, outside of the explored area.

In contrast to the previous setup, the regular curiosity baseline did not learn any useful behaviour---it learned to rotate the joints in a constant direction until reaching joint limits (with the direction of motion varying between different random seeds).
This is most likely caused by more complex exploration required to find the uncertainty in the environment---unlike the previous setup, the box is located farther away from the starting position and it is harder to find during exploration.

As an additional baseline, we used a reaching policy that moves the arm to a random place in the range where the box may be located.

\subsubsection{$\xi$ identification in simulation}
Similarly to previous setups, we train a recurrent model to predict the position of the box given data collected by each baseline policy.
The box position estimation errors (in meters) are presented in Table~\ref{tab:tau_predict}.

We observe that, similarly to previous experiments, the proposed method allows for most accurate estimation of dynamics parameters of all evaluated methods.
The proposed method allows for most accurate estimation of the box position, with an average error of 3 cm.
Other methods either produce an error of 21cm, which is a result of predicting the center of the range.

\subsubsection{State prediction in simulation}
Similarly to previous setups, we also evaluate the model's ability to learn about state transitions in data collected a reference policy, using data collected with each method. 
The reference policy is a policy that reaches the inside of the box (knowing its  ground-truth location).

The results are shown in the third column of Table~\ref{tab:state_predict}.
The prediction error is, at first, the same for all three methods; however, between 50--150 samples into the episode (the time it takes the robot to find the box), it starts to drop for the domain curious approach, reaching a slightly lower final value.
For other approaches, there is no noticeable learning

\subsubsection{Real-world $\xi$ identification}
\begin{figure}
    \centering
    \includegraphics[width=.65\linewidth]{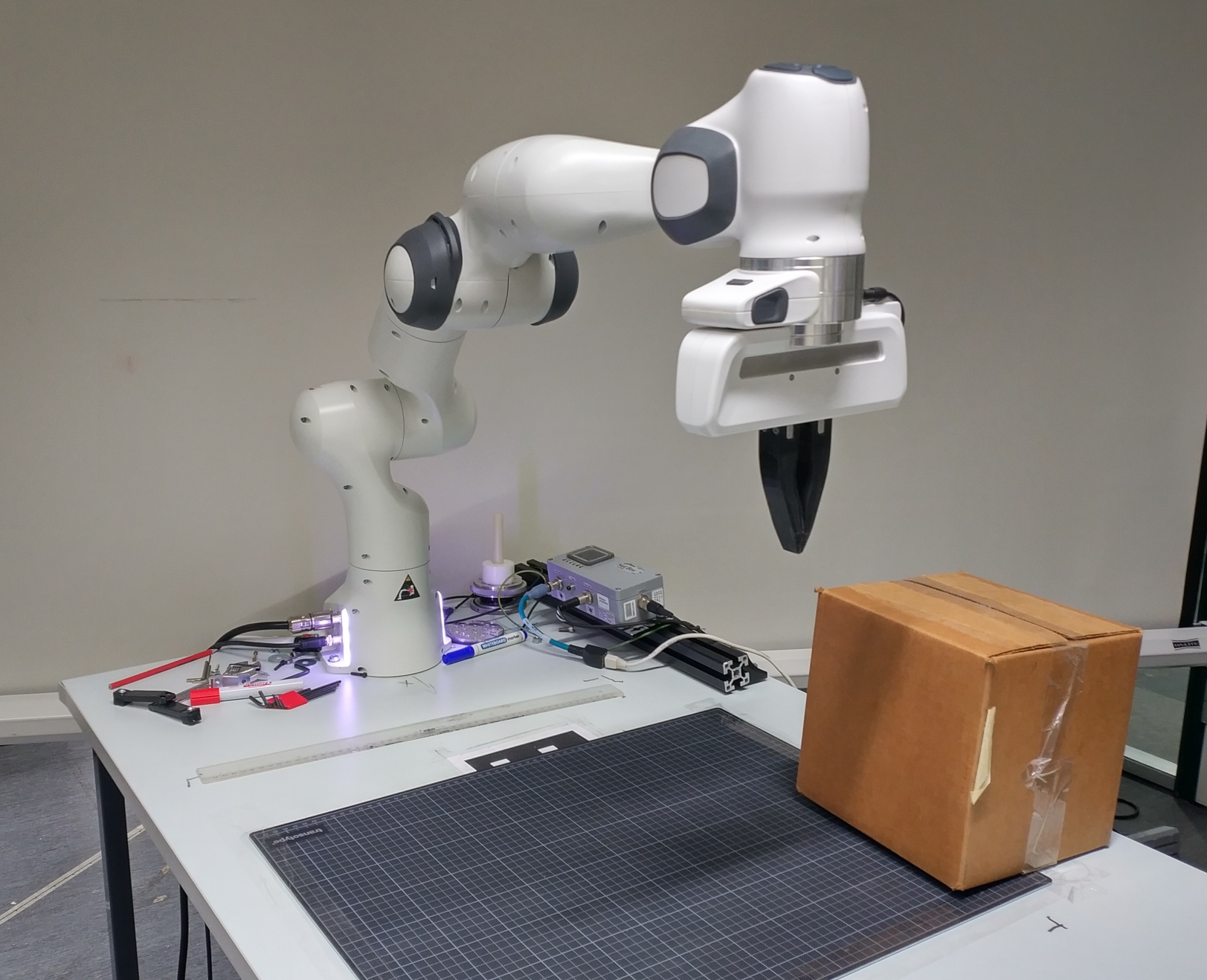}
    \caption{The real world FindBox setup with Franka Panda.}
    \vspace{-5pt}
    \label{fig:rw_setup}
    \vspace{-16pt}
\end{figure}

To verify the method's capability to identify the task variable in real-world conditions, we constructed an analogous setup with a real-world Franka Panda robot arm.
The setup is shown in Figure~\ref{fig:rw_setup}.

In this setup, we have compared the performance of the proposed method to the random reaching baseline.
The baseline was selected because it was the best performing simulated baseline and because---in contrast to the random and curious policies---it is safe to run on physical hardware.
We collected a total 382 trajectories from 26 different box locations for the proposed method and a total of 506 trajectories for the same box positions.
Since the number of rollouts can be different for each evaluation point, each sequence in the loss term was weighted to correct for this dataset imbalance.
Since the real-world table has a width of 80cm, and the size of the box is 20x20cm, we only tested $x$ positions in the range of -30 -- 30cm (to reproduce the range used in simulation, it would be required to have half of the box in the air unsupported by the table, which would likely result in the box being dropped when touched by the robot).
We then split each dataset, leaving 60\% of the trajectories for training and 40\% for the test set.
In order to prevent overfitting, the trained network was heavily regularized---we used weight decay value of $10^{-5}$ and injected random noise with a standard deviation of $10^{-2}$ into the data.
We also used a smaller neural network, with two hidden layers with the size of 16. 

\begin{figure}  
    \centering
    \includegraphics[width=.75\linewidth]{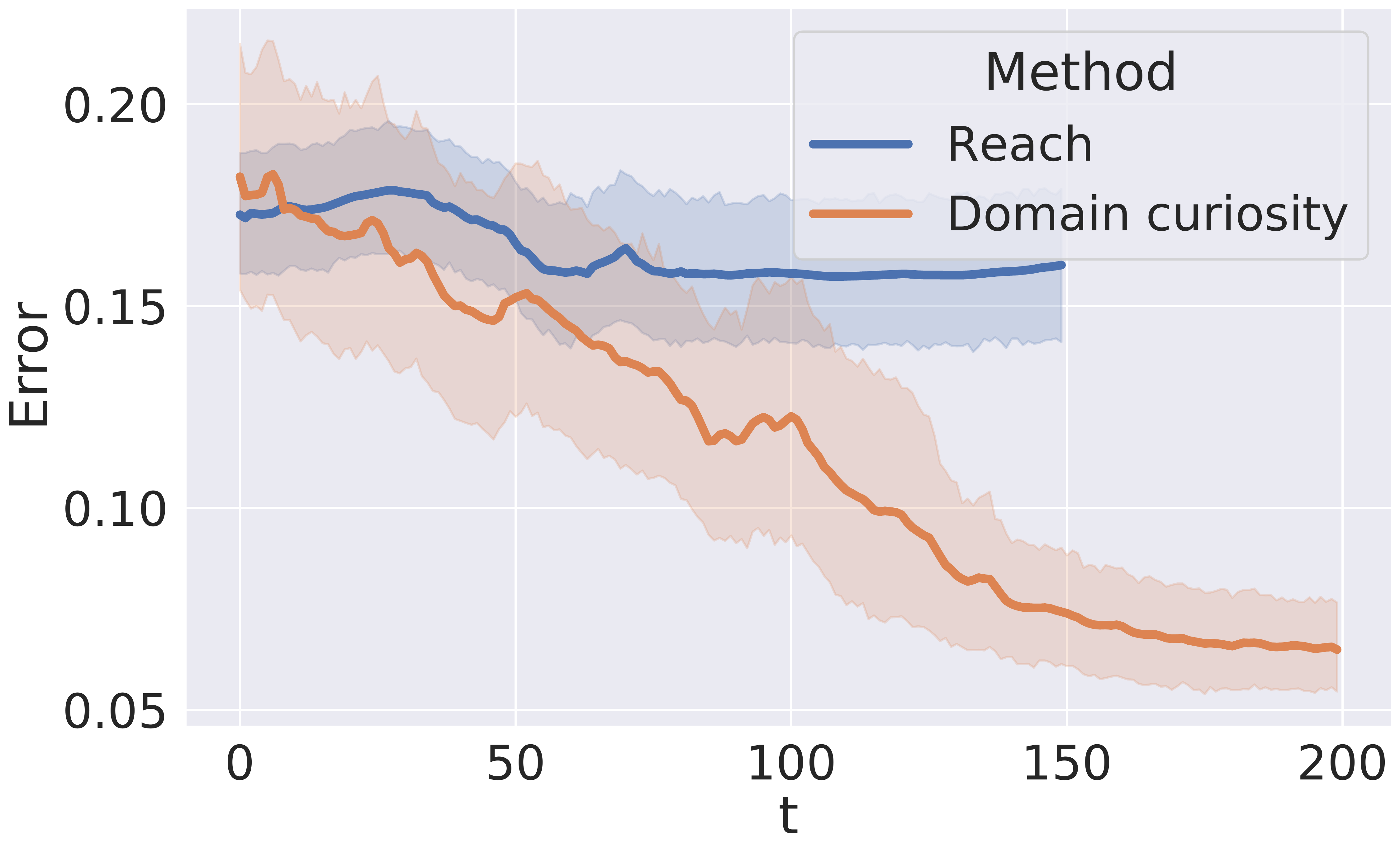}
    \vspace{-5pt}
    \caption{Box position estimation errors $t$ timesteps into an episode for the real-world Box setup with the proposed method and the reaching baseline}
    \label{fig:rw_box_errs}
    \vspace{-14pt}
\end{figure}

The results of this evaluation, in the form of prediction error $t$ timesteps into the episode, are shown in Figure~\ref{fig:rw_box_errs} (where each timestep is 20ms).
Similarly to the simulated results, the proposed approach allows for accurate identification of the location of the box, reaching much smaller position estimation errors than the random reaching baseline.
By analyzing the shape of the domain curious curve, we see that at first, the uncertainty about the position of the box is quite large at around 17cm.
After about 25 timesteps (0.5 seconds), when the robot starts the sweeping motion, the uncertainty starts to fall down, until it finally settles at around 6 cm around 175 (3.5 seconds) timesteps into the episode.
This error is a bit larger than what was observed in simulation (~3cm), but is still a noticeable improvement over the random reaching baseline policy.

For the random reaching policy, the error starts at more or less the same value of 16cm, and, on average, it slightly improves around 40--80 timesteps (0.8--1.6 seconds), which roughly corresponds to the time it takes the robot to reach the goal position.
This indicates that the robot was able to make at least some predictions about the location of the box given the state transitions collected by the reaching policy, but these transitions were far less informative than data collected by the proposed domain curious approach.

\vspace{-4pt}
\section{Conclusions}
\vspace{-2pt}
In this paper, we presented a method for training a \textit{domain curious policy}---a policy specifically optimized to maximize the agent's information gain about the environment.
We demonstrated that the method focuses on the domain-specific, learnable aspects of the environment while ignoring pure noise, unlike standard curiosity-based methods. We also demonstrated that the learned policies can be used to successfully learn about the unknown aspects of the environment, both in simulation and on a real robot setup.

In the proposed method, the policy is only conditioned on the current state, and has no direct measure of how much the agent has already learned.
In simple environments, a successful policy can still be learned by correlating knowledge with other elements of the state space, such as position, with heuristics like \textit{if the agent is moving towards the orange spot, it hasn't visited the spot yet}.
Extending the proposed method to more complex situations may require conditioning the policy on the current task variable and its uncertainty.

To apply the method in multi-task learning scenarios, such as walking to an unknown target, a similarly structured reward prediction model could be used.
The method was studied in the context of dynamics estimation, and it was assumed that only the state transition probabilities depend on $\xi$.
The reward however remains the same across all conditions and it is thus unclear how well the proposed method would perform in such a task.

Ensuring safety is crucial in real-world applications. 
Thus, it is essential that the trained policy explores the environments without putting the agent itself or other actors in the environment at risk~\cite{Huang2019LearningGO}.
While we have provided a simple measure of safety by penalizing contact forces between the robot and other objects in the environment, engineering rewards that assure safety is difficult in general.
While the measure is feasible for cases where acceptable contacts are easy to specify such as pushing, this approach would scale poorly to more complex scenarios, such as compliant interaction between a human and a robot.
One possible way to address this challenge would be to learn safety measures in the simulation, together with the policy and the dynamics model.
This would open a way to increase the efficiency and reliability of sim-to-real transfer by curious domain adaptation, and push sim-to-real methods toward practical applications.

\bibliographystyle{ieeetr}
\bibliography{main}  

\end{document}